\documentclass{article}

\usepackage{arxiv}

\usepackage[utf8]{inputenc} 
\usepackage[T1]{fontenc}    
\usepackage[numbers]{natbib}
\usepackage{hyperref}
\usepackage{url}            
\usepackage{booktabs}       
\usepackage{amsfonts}       
\usepackage{nicefrac}       
\usepackage{microtype}      
\usepackage{lipsum}
\usepackage{graphicx}
\graphicspath{ {./images/} }
\usepackage{times}  
\usepackage{helvet}  
\usepackage{courier}  
\usepackage{graphicx} 
\usepackage{caption} 
\usepackage{algorithm}
\usepackage{amsmath}
\usepackage{algorithmic}
\usepackage{booktabs} 
\usepackage{multirow} 
\usepackage{enumitem}
\usepackage{newfloat}
\usepackage{listings}

\title{EcphoryRAG: Re-Imagining Knowledge-Graph RAG via Human Associative Memory}

\author{
 Zirui Liao \\
  Tsinghua Shenzhen International Graduate School\\
  Shenzhen, China, 518071\\
  \texttt{liao-zr24@mails.tsinghua.edu.cn} \\
}

\begin{document}
\maketitle
\begin{abstract}
Cognitive neuroscience research indicates that humans leverage cues to activate entity-centered memory traces (engrams) for complex, multi-hop recollection. Inspired by this mechanism, we introduce EcphoryRAG, an entity-centric knowledge graph RAG framework. During indexing, EcphoryRAG extracts and stores only core entities with corresponding metadata, a lightweight approach that reduces token consumption by up to 94\% compared to other structured RAG systems. For retrieval, the system first extracts cue entities from queries, then performs a scalable multi-hop associative search across the knowledge graph. Crucially, EcphoryRAG dynamically infers implicit relations between entities to populate context, enabling deep reasoning without exhaustive pre-enumeration of relationships. Extensive evaluations on the 2WikiMultiHop, HotpotQA, and MuSiQue benchmarks demonstrate that EcphoryRAG sets a new state-of-the-art, improving the average Exact Match (EM) score from 0.392 to 0.474 over strong KG-RAG methods like HippoRAG. These results validate the efficacy of the entity-cue-multi-hop retrieval paradigm for complex question answering.
\end{abstract}

\section{Introduction}
The capacity for lifelong acquisition, integration, and reasoning over ever-expanding knowledge is both a defining trait of human cognition and a pivotal challenge for Large Language Models (LLMs). Despite their impressive parametric memory, these models are inherently static and susceptible to hallucination. Retrieval-Augmented Generation (RAG) has consequently become the de-facto paradigm for grounding LLMs in external, verifiable sources\citep{rag}.

The evolution of RAG has progressed from simple dense retrieval towards more structured approaches. For example, methods like ReAct \citep{Yao2023ReAct} simulate multi-hop reasoning through iterative LLM calls, but this approach can introduce high latency and propagate errors.  Knowledge Graph-Augmented RAG (KG-RAG) offers a more robust solution by explicitly modeling relationships \citep{graphrag}.However, current KG-RAG implementations present a difficult trade-off. Static, pre-computed graphs offer efficient querying but are costly to update and struggle with unfamiliar queries. In contrast, dynamic graph traversal provides flexibility but often depends on slow, token-by-token LLM guidance for pathfinding \citep{Zhang2024ThinkonGraph}. There is a clear need for a framework that merges the structural efficiency of static graphs with the adaptability of dynamic methods.

\begin{figure}[t]
\centering
\includegraphics[width=\columnwidth]{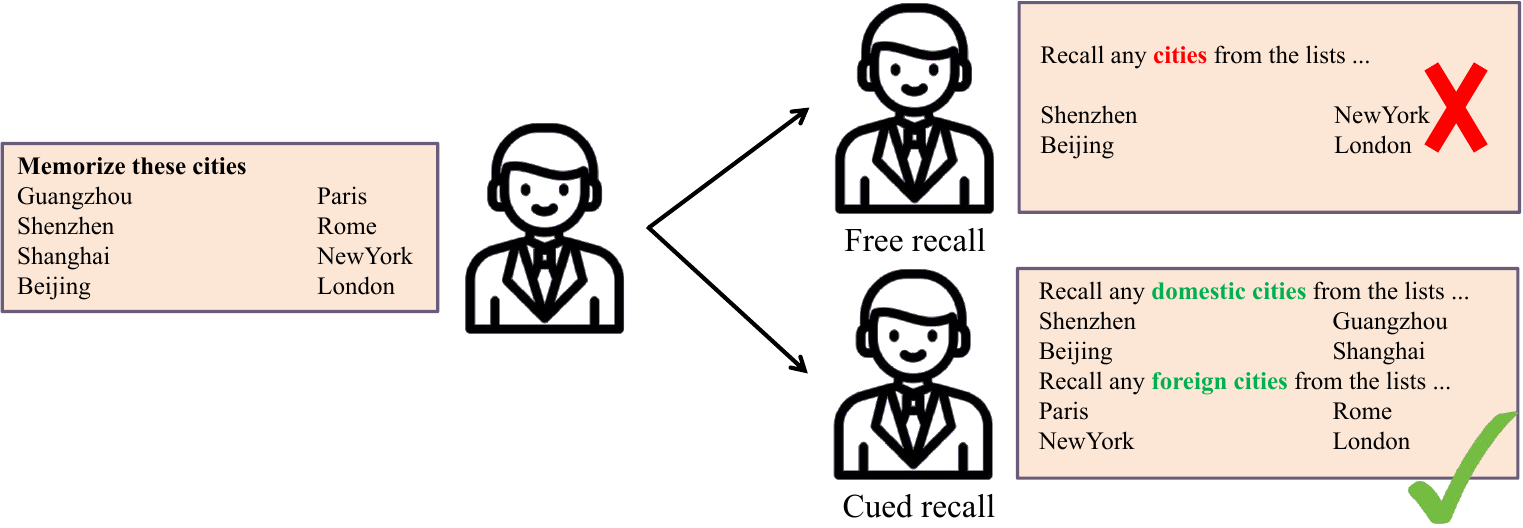} 
\caption{The cognitive principle of cued recall that inspires our work. In a classic memory experiment, a general prompt (\textbf{Free Recall}) often leads to incomplete retrieval. However, specific \textbf{Cues} (e.g., "domestic" or "foreign") activate targeted memory traces, enabling complete and accurate recall. EcphoryRAG is designed to operationalize this powerful principle for complex question answering.}
\label{fig:cued_recall_concept}
\end{figure}

Cognitive science offers a different model for knowledge retrieval. Human memory, for instance, does not perform an exhaustive search; recall is instead an efficient, cue-driven process. As shown in Figure~\ref{fig:cued_recall_concept}, a general prompt may yield incomplete information, whereas specific \textbf{cues} can activate targeted memory traces—or \textbf{engrams}\footnote{We use \textit{engram} computationally to denote a structured record (an entity plus its metadata), drawing an analogy to the biological ensemble of neurons without claiming a direct equivalence.}—for precise retrieval \citep{Tulving_1984, engram}. The principle by which a partial cue triggers the reconstruction of a full memory is known as \textbf{ecphory}. Although recent systems like HippoRAG2\citep{hipporag2} draw inspiration from human memory, they do not fully replicate this dynamic, cue-based mechanism. Therefore, a practical and scalable computational model of ecphory for RAG has yet to be developed.

\begin{figure*}[t]
\centering
\includegraphics[width=0.8\textwidth]{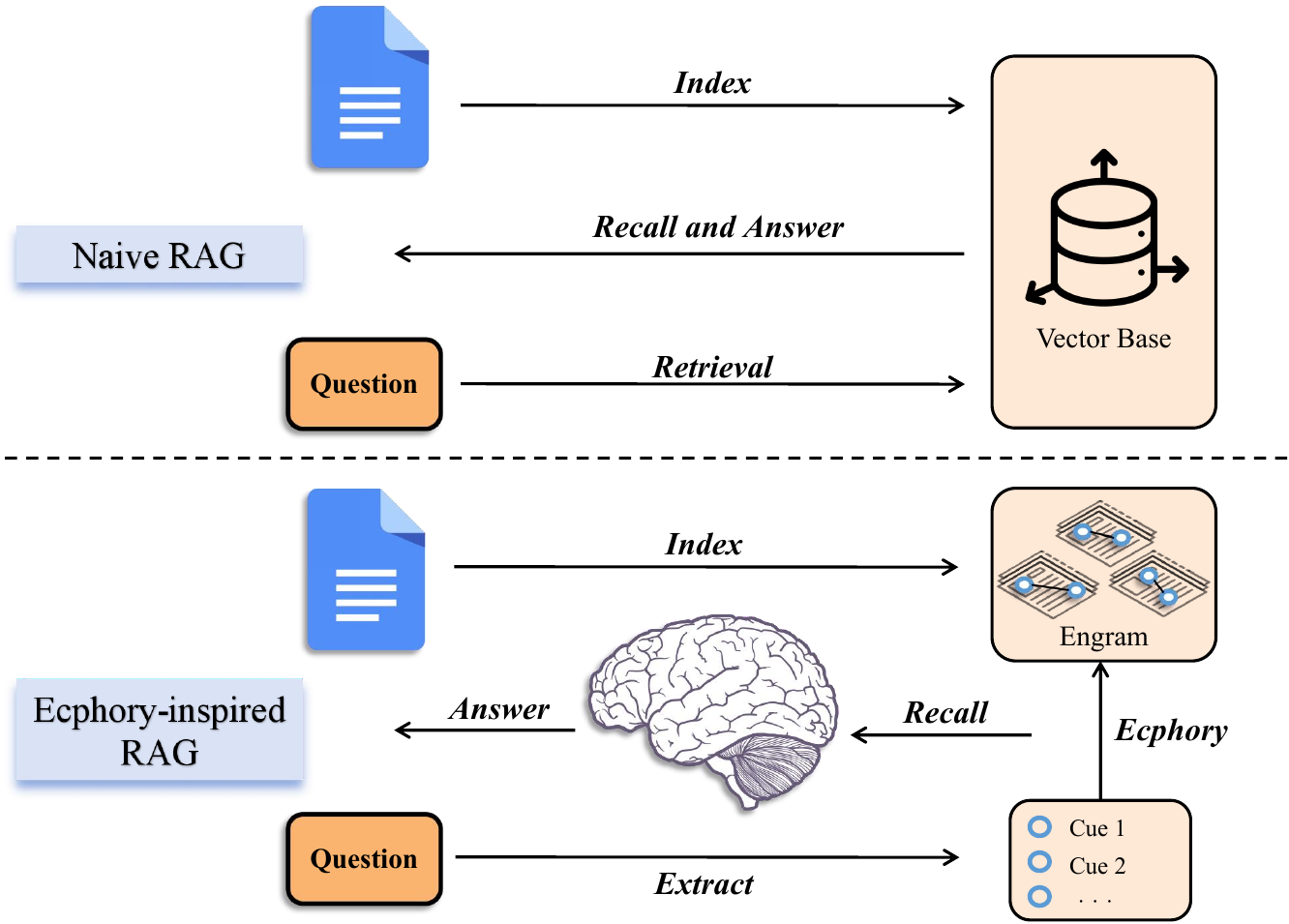}
\caption{A comparison of retrieval paradigms. \textbf{(Top) Naive RAG}: A question directly retrieves information from a monolithic vector base using a single semantic search step. This process is simple but often fails to connect disparate facts. \textbf{(Bottom) Ecphory-inspired RAG}: Our proposed workflow, which mimics human memory. A question is first processed to \textit{Extract} specific Cues. These cues then trigger a targeted \textit{Recall} process (Ecphory) from a structured knowledge base of \textit{Engrams}. The retrieved, relevant information is then synthesized by an LLM (the "brain") to produce the final, reasoned answer.}
\label{fig:framework_comparison}
\end{figure*}

To address this gap, we introduce \textbf{EcphoryRAG}, a novel framework that instantiates the principle of ecphory for knowledge reasoning. As illustrated in Figure~\ref{fig:framework_comparison}, EcphoryRAG fundamentally redesigns the standard RAG workflow. Instead of direct, monolithic retrieval, our framework first uses an LLM to \textit{Extract} specific Cues from the user's query. These cues then trigger a targeted \textit{Recall} process (Ecphory) from a structured knowledge base of engrams. This design achieves the structural integrity of KG-RAG while maintaining the adaptability of dynamic methods, all inspired by the efficiency of human memory.

Our primary contributions are:
\begin{enumerate}[label=(\arabic*)]
    \item We propose EcphoryRAG, a new RAG framework that operationalizes the cognitive principle of ecphory through a cue-driven, multi-hop retrieval mechanism designed for complex reasoning.
    \item We introduce a hybrid associative search algorithm that combines explicit graph traversal with implicit semantic expansion, enabling the discovery of complex, latent reasoning paths that are missed by other methods.
    \item We demonstrate through extensive experiments that EcphoryRAG not only establishes a new state-of-the-art on three challenging multi-hop QA benchmarks, but also achieves this with remarkable efficiency, reducing offline indexing token costs by up to 18x compared to other structured RAG systems.
\end{enumerate}

\begin{figure*}[t!]
\centering
\includegraphics[width=0.9\textwidth]{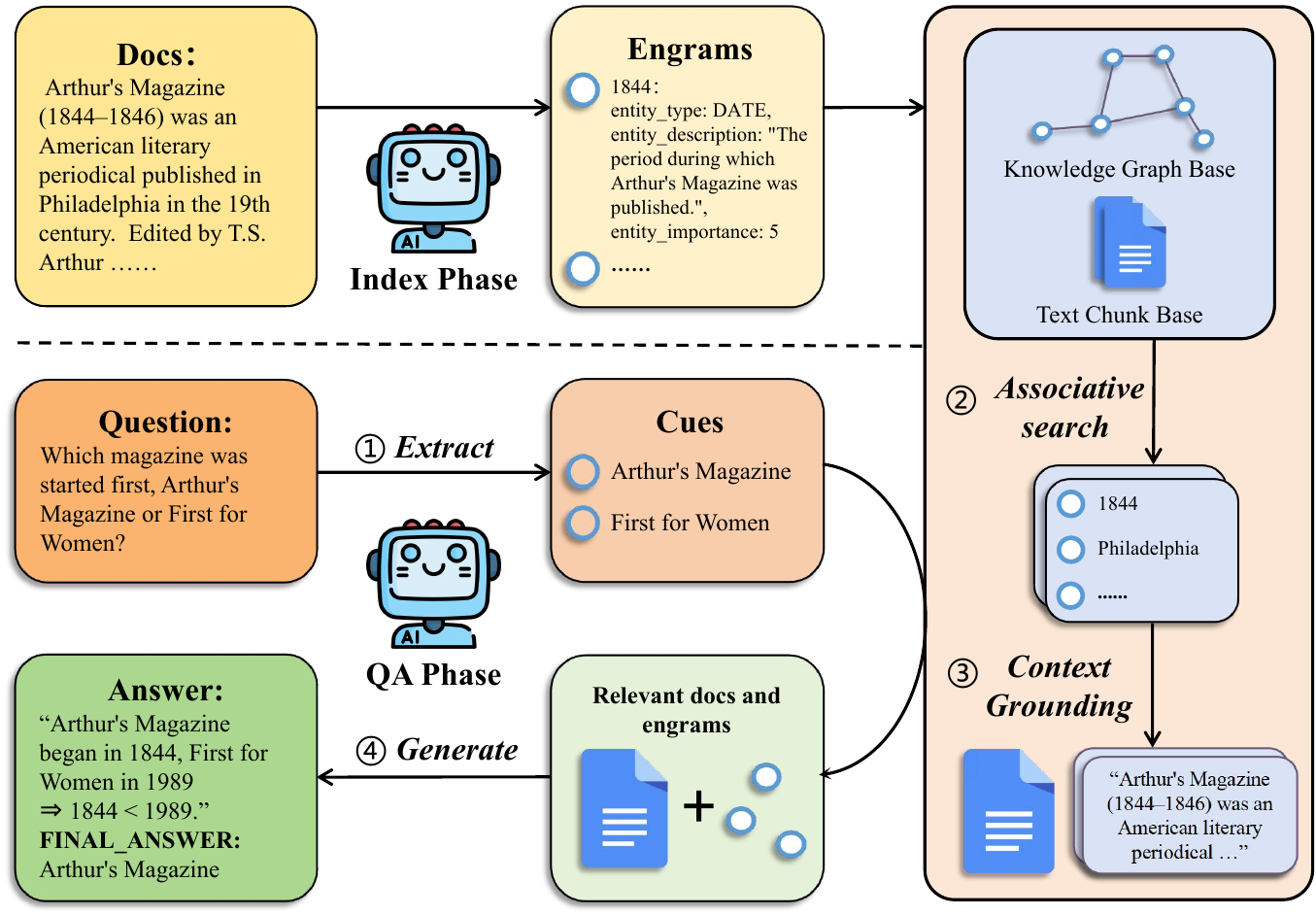}
\caption{The detailed end-to-end workflow of EcphoryRAG, separated into an offline Index Phase and an online QA Phase.
\textbf{Index Phase (top):} Raw documents are processed by an LLM to extract structured \textbf{Engrams} (e.g., the entity '1844' with its type, description, importance score and so on). These engrams are stored in a multi-granular memory system, comprising a \textbf{Knowledge Graph Base} for relational structure and a \textbf{Text Chunk Base} for source evidence.
\textbf{QA Phase (bottom):} The process begins with a user's \textit{Question}. \textbf{(1) Cue Extraction:} An LLM identifies key entities to serve as retrieval Cues. \textbf{(2) Associative Search:} These cues initiate a multi-hop search across the knowledge graph, activating related engrams. \textbf{(3) Context Grounding:} The system retrieves the original text chunks associated with the activated engrams. \textbf{(4) Generation:} The retrieved engrams and their grounded text are combined into a rich context, which the LLM uses to perform step-by-step reasoning and generate the final answer.}
\label{fig:detailed_workflow}
\end{figure*}


\section{Related Work}
Our work builds upon advancements in Retrieval-Augmented Generation (RAG), drawing from the evolution of structured retrieval and the emerging field of cognitive-inspired memory systems.

\subsection{Evolution of RAG: From Dense to Structured}
Early RAG systems, such as Dense Passage Retrieval (DPR) \citep{Karpukhin2020Dense}, primarily used dense vectors to retrieve independent text passages. While effective for simple factoid queries, this approach struggles with multi-hop reasoning, as it lacks an understanding of the relational context between facts. To address this, iterative RAG methods like ReAct \citep{Yao2023ReAct} were developed to decompose complex questions into a series of retrieval steps. However, this often incurs significant latency and token costs due to multiple LLM calls and still lacks an explicit, structured memory to guide the multi-step retrieval process efficiently.

\subsection{Knowledge Graph-Augmented RAG (KG-RAG)}
To incorporate explicit structure, KG-RAG systems emerged, but they face a critical trade-off between static indexing and dynamic traversal. On one hand, static methods like QA-GNN \citep{Yasunaga2021QAGNN} and HippoRAG 2 \citep{hipporag2} pre-compute a large knowledge graph, enabling efficient online retrieval but suffering from high update costs and inflexibility. In contrast, dynamic methods like Think-on-Graph \citep{Zhang2024ThinkonGraph} traverse the knowledge space on-the-fly, offering greater flexibility but often relying on inefficient, LLM-guided navigation. EcphoryRAG addresses this trade-off by using a lightweight, pre-built entity graph but performing a highly dynamic, cue-driven traversal that avoids costly token-by-token LLM guidance.

\subsection{Cognitive and Memory-Inspired RAG}
A growing body of research draws inspiration from cognitive science to design more robust RAG systems. Human memory retrieval is understood as a cue-driven, reconstructive process known as \textbf{ecphory}, where a partial cue activates a complete memory representation, or engram \citep{Tulving_1984, Kruschke2022Neuro}. While recent frameworks like HippoRAG \citep{gutierrez2024hipporag} have adopted this neurobiological perspective, their retrieval is often initiated by simple named entity cues. EcphoryRAG advances this paradigm by operationalizing a more nuanced model of ecphory, using the entire query as a rich cue to dynamically activate and explore a relevant subgraph of engrams, more closely mimicking the flexible, context-dependent nature of human memory recall.


\section{The EcphoryRAG Framework}

This section details our proposed framework, EcphoryRAG, which is inspired by the cognitive process of memory ecphory—the mechanism by which cues trigger the retrieval of stored memory traces \citep{Tulving_1984}. Drawing an analogy from human memory, EcphoryRAG computationalizes this process for complex, multi-hop question answering. It comprises two main phases: an offline \textbf{Memory System Construction} phase to index knowledge, and an online \textbf{Retrieval as Ecphory} phase that simulates cue-driven recall. The detailed workflow is illustrated in Figure~\ref{fig:detailed_workflow}.

\subsection{Memory System Construction (Indexing)}
The goal of this offline phase is to transform an unstructured document corpus into a structured and interconnected memory system \(\mathcal{M}\).

\subsubsection{Notation.} 
Throughout this section, we denote a user query as \(\mathcal{Q}\) and a document corpus as \(\mathcal{D}_{\text{raw}}\). Our constructed knowledge graph is \(\mathcal{G} = (\mathcal{V}, \mathcal{E'})\), where \(\mathcal{V}\) is the set of entity nodes and \(\mathcal{E'}\) is the set of edges. An individual entity is denoted by \(e\), and its vector embedding by \(\mathbf{v}_e\).

\subsubsection{Entities as Knowledge Engrams.}
We posit that entities are the fundamental anchors of knowledge, or engrams. For each document \(d \in \mathcal{D}_{\text{raw}}\), we first segment it into text chunks \(\mathcal{C}(d)\). We then employ a Large Language Model (LLM) to extract a set of entities \(\mathcal{E}(c_j)\) from each chunk \(c_j\):
\begin{equation}
\mathcal{E}(c_j) = \mathrm{LLM}_{\mathrm{extract}}(c_j, \Pi_{\mathrm{entity}})
\label{eq:entity_extraction}
\end{equation}
where \(\Pi_{\mathrm{entity}}\) is a prompt template for structured entity extraction. Each entity \(e \in \mathcal{E}(c_j)\) is stored as a tuple with its core metadata: \(e = (\mathrm{id}_e, \mathrm{name}_e, \mathrm{type}_e, \mathrm{desc}_e, \mathrm{src}_e)\), where \(\mathrm{src}_e\) is a pointer to its source chunk.

\subsubsection{Building the Associative Knowledge Graph.}
To model the associative links between engrams, we construct a knowledge graph \(\mathcal{G}\). The nodes \(\mathcal{V}\) are the extracted entities \(\mathcal{E}_{\text{all}}\). Edges \(\mathcal{E'}\) represent co-occurrence relationships: an edge \((e_i, e_j)\) exists if entities \(e_i\) and \(e_j\) appear in the same text chunk. The graph is treated as \textbf{undirected and unweighted} to model symmetric associations, forming the structural backbone for our associative search.

\subsubsection{Multi-Granularity Vector Indexing.}
We create separate ANN indices for entities and chunks.
\begin{itemize}
    \item \textbf{Entity Index (\(I_E\))}: For each entity \(e\), we compute its embedding \(\mathbf{v}_e\) from its name and description:
    \begin{equation}
    \mathbf{v}_e = \Phi_{\mathrm{emb}}(\mathrm{name}_e \oplus \mathrm{desc}_e)
    \end{equation}
    where \(\oplus\) denotes concatenation.
    \item \textbf{Chunk Index (\(I_C\))}: We also compute and index an embedding \(\mathbf{v}_c\) for each source text chunk \(c\).
\end{itemize}
The complete memory system is thus a composite structure \(\mathcal{M} = (\mathcal{E}_{\mathrm{all}}, \mathcal{C}_{\mathrm{all}}, \mathcal{G}, I_E, I_C)\).

\subsection{Retrieval as Ecphory: A Multi-Hop Associative Search}

When a question \(\mathcal{Q}\) is received, EcphoryRAG simulates memory recall through a multi-stage retrieval process designed to gather a rich and relevant context. This process involves initial activation, associative expansion, and final re-ranking.

\subsubsection{Initial Context Activation.}
This first step aims to retrieve a broad set of potentially relevant memory traces (both entities and text chunks) based on the user's query.

\begin{enumerate}[label=(\arabic*)]
    \item \textbf{Primary Engram Retrieval:} We first compute a dense vector embedding \(\mathbf{v}_{\mathcal{Q}}\) for the entire query text. This embedding is used to search the entity index \(I_E\), activating an initial set of engrams (entities) that are semantically closest to the query's intent.
    \begin{equation}
        \mathcal{E}_{\text{init}} = \mathrm{Search}(I_E, \mathbf{v}_{\mathcal{Q}}, k_{\text{initial}})
    \end{equation}
    where \(k_{\text{initial}}\) is a hyperparameter (e.g., 20) defining the breadth of the initial search.

    \item \textbf{Parallel Chunk Retrieval:} Simultaneously, we perform two types of text chunk retrieval. First, we perform \textbf{context grounding} by fetching the source chunks \(\mathcal{C}_{\text{assoc}}\) associated with the initially retrieved entities \(\mathcal{E}_{\text{init}}\). Second, we conduct a \textbf{direct search} in the chunk index \(I_C\) using the query embedding \(\mathbf{v}_{\mathcal{Q}}\) to find semantically relevant chunks \(\mathcal{C}_{\text{direct}}\). These sets are merged and deduplicated to form the initial context \(\mathcal{C}_{\text{init}}\).
\end{enumerate}

\subsubsection{Associative Search: Multi-Depth Expansion.}
To uncover deeper, multi-hop connections, we perform an iterative associative search for \(\text{retrieval\_depth}\) rounds, starting with the initially retrieved entities \(\mathcal{E}_{\text{init}}\). In each round \(l\):
\begin{enumerate}[label=(\arabic*)]
    \item \textbf{Seed Selection:} We select a set of high-confidence "seed" entities \(\mathcal{E}_{\text{seed}}^{(l-1)}\) from the entities retrieved so far. These are typically the entities with the highest similarity scores to the query.

    \item \textbf{Weighted Centroid Embedding:} Instead of a simple mean, we compute a weighted average embedding of the seed entities. The weight for each seed entity is proportional to its initial similarity score, allowing more relevant entities to have a greater influence on the search direction.
    \begin{equation}
        \bar{\mathbf{v}}_{\text{weighted}}^{(l-1)} = \sum_{e \in \mathcal{E}_{\text{seed}}^{(l-1)}} w_e \cdot \mathbf{v}_e
    \end{equation}
    where \(w_e = \mathrm{sim}(e, \mathcal{Q}) / \sum_{j} \mathrm{sim}(e_j, \mathcal{Q})\).
    
    \item \textbf{Expansion and Fusion:} This weighted centroid embedding is then used to perform a new search in the entity index \(I_E\), retrieving a new set of associated entities \(\mathcal{E}_{\text{assoc}}^{(l)}\). These new entities are deduplicated and fused with the existing set of retrieved entities.
\end{enumerate}
This process allows the search to "walk" through the embedding space, following a path guided by the most relevant concepts discovered at each step.

\subsubsection{Final Re-ranking and Selection.}
After the associative search completes, the expanded set of all retrieved entities, \(\mathcal{E}_{\text{all}}\), undergoes a final, crucial re-ranking step. We re-calculate the cosine similarity of every entity in \(\mathcal{E}_{\text{all}}\) directly against the original query embedding \(\mathbf{v}_{\mathcal{Q}}\). This ensures that the final ranking is anchored to the user's original intent, mitigating topic drift that can occur during multi-hop expansion.
\begin{equation}
    \mathrm{score}(e) = \frac{\mathbf{v}_e \cdot \mathbf{v}_{\mathcal{Q}}}{\|\mathbf{v}_e\| \|\mathbf{v}_{\mathcal{Q}}\|}
\end{equation}
The entities are then sorted by this new score, and the top-\(k_{\text{final}}\) entities are selected as the final reasoning path \(\mathcal{E}_{\text{final}}\). The initial context chunks \(\mathcal{C}_{\text{init}}\) and the final entities \(\mathcal{E}_{\text{final}}\) are then passed to the generator LLM. The entire online workflow is summarized in Algorithm~\ref{alg:ecphoryrag}.

\begin{algorithm}[tb]
\caption{The EcphoryRAG Online Workflow}
\label{alg:ecphoryrag}
\textbf{Input}: Query \(\mathcal{Q}\), Memory System \(\mathcal{M}=(\mathcal{G}, I_E, I_C)\) \\
\textbf{Parameters}: \(k_{\text{initial}}\), \(\text{retrieval\_depth}\), \(k_{\text{final}}\) \\
\textbf{Output}: Answer \(\mathcal{A}\), Provenance \(\mathcal{P}\)

\begin{algorithmic}[1] 
\STATE \(\mathbf{v}_{\mathcal{Q}} \leftarrow \mathrm{Embed}(\mathcal{Q})\)
\STATE \COMMENT{--- Initial Context Activation ---}
\STATE \(\mathcal{E}_{\text{init}} \leftarrow \mathrm{Search}(I_E, \mathbf{v}_{\mathcal{Q}}, k_{\text{initial}})\)
\STATE \(\mathcal{C}_{\text{assoc}} \leftarrow \mathrm{GetSourceChunks}(\mathcal{E}_{\text{init}})\)
\STATE \(\mathcal{C}_{\text{direct}} \leftarrow \mathrm{Search}(I_C, \mathbf{v}_{\mathcal{Q}}, 10)\)
\STATE \(\mathcal{C}_{\text{init}} \leftarrow \mathrm{Deduplicate}(\mathcal{C}_{\text{assoc}} \cup \mathcal{C}_{\text{direct}})\)
\STATE \(\mathcal{E}_{\text{all}} \leftarrow \mathcal{E}_{\text{init}}\)
\STATE
\STATE \COMMENT{--- Associative Search ---}
\IF{\(\text{retrieval\_depth} > 0\)}
    \FOR{\(l = 1\) \TO \(\text{retrieval\_depth}\)} 
        \STATE \(\mathcal{E}_{\text{seed}} \leftarrow \mathrm{SelectSeeds}(\mathcal{E}_{\text{all}})\)
        \STATE \(\bar{\mathbf{v}}_{\text{weighted}} \leftarrow \mathrm{WeightedCentroid}(\mathcal{E}_{\text{seed}})\)
        \STATE \(\mathcal{E}_{\text{new}} \leftarrow \mathrm{Search}(I_E, \bar{\mathbf{v}}_{\text{weighted}}, k_{\text{initial}} * 3)\)
        \STATE \(\mathcal{E}_{\text{all}} \leftarrow \mathrm{Deduplicate}(\mathcal{E}_{\text{all}} \cup \mathcal{E}_{\text{new}})\)
    \ENDFOR
\ENDIF
\STATE
\STATE \COMMENT{--- Final Re-ranking and Generation ---}
\STATE \(\mathcal{E}_{\text{ranked}} \leftarrow \mathrm{RerankWithQuery}(\mathcal{E}_{\text{all}}, \mathbf{v}_{\mathcal{Q}})\)
\STATE \(\mathcal{E}_{\text{final}} \leftarrow \mathcal{E}_{\text{ranked}}[:k_{\text{final}}]\)
\STATE \(\mathcal{C}_{\text{final}} \leftarrow \mathcal{C}_{\text{init}}[:5]\) \COMMENT{Select top initial chunks}
\STATE \(\Pi_{\text{gen}} \leftarrow \mathrm{FormatPrompt}(\mathcal{Q}, \mathcal{E}_{\text{final}}, \mathcal{C}_{\text{final}})\)
\STATE \(\mathcal{A} \leftarrow \mathrm{LLM}_{\mathrm{gen}}(\Pi_{\text{gen}})\)
\STATE \(\mathcal{P} \leftarrow (\mathcal{E}_{\text{final}}, \mathcal{C}_{\text{final}})\)
\STATE \textbf{return} \(\mathcal{A}, \mathcal{P}\)
\end{algorithmic}
\end{algorithm}



\section{Experiments}
\begin{table*}[t]
\centering
\caption{Comprehensive performance and efficiency comparison on multi-hop QA benchmarks. Results for EcphoryRAG are mean ± std. dev. over 10 runs. Baselines were run once. Bold and underline indicate the best and second-best results, respectively. Efficiency metrics IT and QT are averaged over all datasets.}
\label{tab:main_results_combined}
\resizebox{\textwidth}{!}{%
\begin{tabular}{@{}lccccccccrr@{}}
\toprule
\multirow{2}{*}{\textbf{Method}} & \multicolumn{2}{c}{\textbf{2WikiMultiHop}} & \multicolumn{2}{c}{\textbf{HotpotQA}} & \multicolumn{2}{c}{\textbf{MuSiQue}} & \multicolumn{2}{c}{\textbf{Average Performance}} & \multicolumn{2}{c}{\textbf{Average Efficiency}} \\
\cmidrule(lr){2-3} \cmidrule(lr){4-5} \cmidrule(lr){6-7} \cmidrule(lr){8-9} \cmidrule(lr){10-11}
 & EM & F1 & EM & F1 & EM & F1 & EM & F1 & IT (Tokens) & QT (Tokens) \\
\midrule
Vanilla RAG & 0.360 & 0.381 & 0.284 & 0.325 & 0.170 & 0.231 & 0.271 & 0.312 & \textbf{11.2k} & 848.1k \\
LightRAG & 0.130 & 0.141 & 0.210 & 0.233 & 0.045 & 0.090 & 0.128 & 0.155 & 36.4M & \textbf{462.9k} \\
HippoRAG2 & \underline{0.404} & \textbf{0.520} & \underline{0.580} & \underline{0.716} & \underline{0.186} & \underline{0.362} & \underline{0.390} & \underline{0.533} & 6.6M & 832.5k \\
\textbf{EcphoryRAG} & \textbf{0.406±.004} & \underline{0.454±.005} & \textbf{0.722±.006} & \textbf{0.814±.004} & \textbf{0.295±.005} & \textbf{0.369±.006}  &  \textbf{0.475} & \textbf{0.547} & \underline{2.0M} & 1.3M \\
\bottomrule
\end{tabular}%
}
\end{table*}
We conduct a series of experiments to comprehensively evaluate EcphoryRAG, focusing on its multi-hop reasoning performance, computational efficiency, and the impact of its core architectural components.

\subsection{Experimental Setup}

\subsubsection{Datasets.}
To rigorously test our framework, we selected three widely-used multi-hop QA benchmarks, each presenting a unique reasoning challenge:
\begin{itemize}
    \item \textbf{2WikiMultiHopQA} \citep{2wikiqa}: Requires connecting facts across two distinct Wikipedia articles, testing fundamental 2-hop reasoning.
    \item \textbf{HotpotQA} \citep{hotpotqa}: Demands finding and synthesizing information from multiple supporting documents, emphasizing the model's ability to handle noise and perform information fusion.
    \item \textbf{MuSiQue} \citep{musique}: Features complex questions requiring deeper 2-to-4 reasoning hops, designed to challenge the logical depth of a system.
\end{itemize}
For each dataset, we use a random sample of 500 questions for evaluation.

\subsubsection{Baselines.}
We compare EcphoryRAG against a spectrum of representative RAG models. All baselines are implemented using the default configurations specified in their respective papers and official repositories, unless otherwise noted:
\begin{enumerate}[label=(\arabic*)]
\item \textbf{Vanilla RAG} \citep{rag}: A standard dense retrieval baseline that retrieves text chunks based on vector similarity. It serves to establish the performance floor without structured reasoning. 
\item \textbf{LightRAG} \citep{guo2025lightragsimplefastretrievalaugmented}: A system that integrates a graph index and a dual-level retrieval mechanism, aiming for efficiency. 
\item \textbf{HippoRAG2} \citep{hipporag2}: A prominent neurobiologically-inspired framework using a KG and Personalized PageRank, representing the state-of-the-art in single-step graph-based retrieval.
\end{enumerate}
For the Vanilla RAG model, we tested several values of the topk hyperparameter and selected the topk value that achieved the highest performance.

\subsubsection{Metrics.}
We assess performance using two standard QA metrics and two efficiency metrics:
\begin{itemize}
    \item \textbf{Exact Match (EM)}: The percentage of generated answers that are identical to the ground truth.
    \item \textbf{F1 Score}: The word-level harmonic mean of precision and recall between generated and ground truth answers, robust to minor phrasing differences.
    \item \textbf{Indexing Tokens (IT)}: The total number of tokens consumed by the LLM during the offline index construction phase, measuring the one-time setup cost.
    \item \textbf{Querying Tokens (QT)}: The average number of tokens consumed by the LLM to process a single query, measuring the online operational cost.
\end{itemize}

\subsubsection{Implementation Details.}
\label{sec:implementation_details}
To ensure the robustness of our primary claims, all results for our proposed method, \textbf{EcphoryRAG, are reported as the mean and standard deviation over 10 runs} with different random seeds. Due to computational constraints, all baseline models were evaluated once using a fixed random seed of 42.

\noindent\textbf{Core Components:} The LLM used for both entity extraction and answer generation is \textbf{Phi4} \citep{phi4}, accessed via Ollama. The embedding model is \textbf{bge-m3} \citep{bge-m3}. For EcphoryRAG, the retrieval depth \(L\) is set to 2, and the final context size \(k_{final}\) is tuned for each dataset as described in our ablation studies. A detailed list of all hyperparameters for all models is provided in the Appendix.

\noindent\textbf{Infrastructure and Software:} Experiments were conducted on a server equipped with an NVIDIA RTX 4090 GPU (24GB VRAM), running Ubuntu 22.04. The implementation is built on Python 3.10. Key libraries include numpy (1.26.4), faiss-cpu (1.11.0) for vector indexing, networkx(3.4.2) for graph operations, and langchain(0.3.25) for text processing. The full list of dependencies is available in the supplementary material.

\section{Results and Analysis}

\subsection{Main Performance and Efficiency}
Table~\ref{tab:main_results_combined} consolidates our main findings, comparing both the QA performance and the average computational efficiency of all methods. The results unequivocally demonstrate that EcphoryRAG not only sets a new state-of-the-art in reasoning accuracy but also strikes a highly practical and superior balance of costs.

\subsubsection{Dominant QA Performance.}
EcphoryRAG achieves the highest average EM (0.475) and F1 (0.547) scores, consistently outperforming all baselines. The performance gains are particularly stark when compared to other structured RAG methods. For instance, on the challenging HotpotQA dataset, EcphoryRAG improves the EM score by a remarkable 24.5\% relative to the strong HippoRAG baseline. We performed a paired t-test and found the performance improvements of EcphoryRAG over HippoRAG to be statistically significant (\(p < 0.01\)) across all datasets. This consistent and robust superiority validates that our cognitively inspired design enables more precise and reliable multi-hop reasoning.

\subsubsection{A Superior Efficiency Profile.}
The efficiency results in Table~\ref{tab:main_results_combined} highlight critical architectural trade-offs in modern RAG systems.
\begin{enumerate}[label=(\arabic*)]
    \item \textbf{Highly Efficient Indexing:} EcphoryRAG's key advantage lies in its indexing efficiency. Its average indexing cost (IT) of 2.0M tokens is dramatically lower than other KG-RAG methods—\textbf{3.3x lower} than HippoRAG and a staggering \textbf{18x lower} than LightRAG. This efficiency stems from our streamlined, single-pass entity extraction process, which avoids the costly multi-round retry mechanisms employed by other complex methods to ensure entity recall. This makes EcphoryRAG far more scalable and cost-effective for large, dynamic knowledge bases.

    \item \textbf{Strategic Query-Time Investment:} While EcphoryRAG's average query cost (QT) is higher, this is a conscious design choice. It represents a strategic investment in reasoning quality, where computational resources are allocated to the flexible online retrieval process rather than the rigid, expensive offline indexing phase. The additional query tokens are used to power the deep reasoning over the rich, structured context assembled by our ecphory mechanism. This investment directly fuels the state-of-the-art QA performance, proving to be a highly effective trade-off. A more granular, per-dataset breakdown of these costs is provided in the Appendix.
\end{enumerate}

\subsection{Ablation Studies}
To dissect the contributions of EcphoryRAG's core components, we performed a series of ablation studies. We summarize the key findings here; detailed analyses are provided in the Appendix.

\subsubsection{Ablation on Context Grounding.}
\begin{figure}[t]         
\centering
\begin{minipage}[t]{0.48\linewidth}
  \centering
  \includegraphics[width=\linewidth]{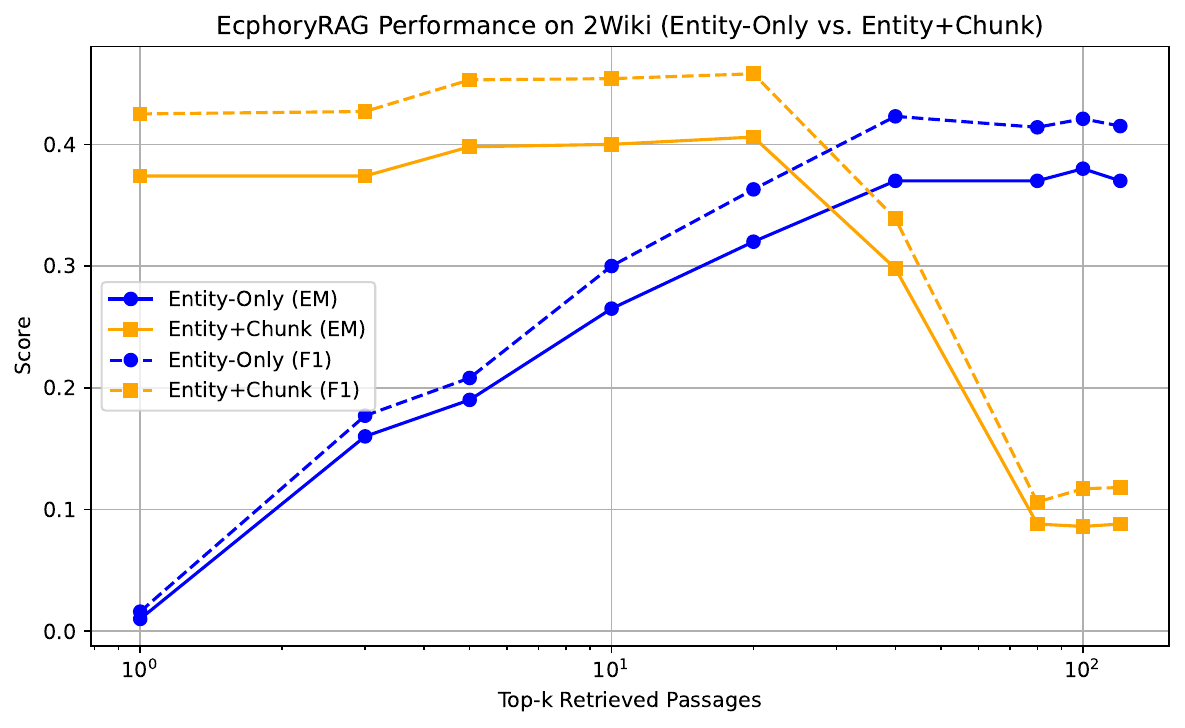}
  \caption{Ablation on context components on the 2Wiki dataset. The full ``Entity+Chunk'' method (orange) vastly outperforms the ``Entity-Only'' approach (blue), though it is more sensitive to noise at larger values of~$k$.}
  \label{fig:context_ablation}
\end{minipage}\hfill
\begin{minipage}[t]{0.48\linewidth}
  \centering
  \includegraphics[width=\linewidth]{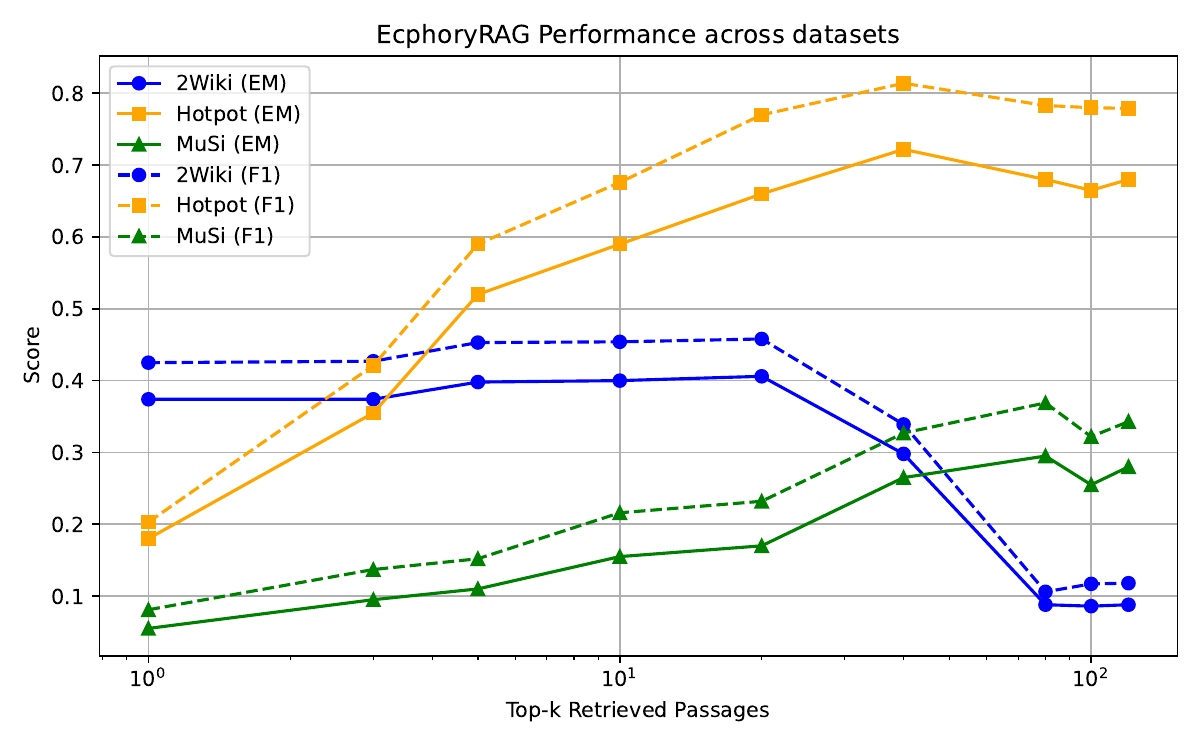}
  \caption{Performance of EcphoryRAG with varying numbers of retrieved passages~($k$). The optimal value of~$k$ is dataset-dependent, reflecting differences in evidence dispersion and noise sensitivity.}
  \label{fig:k_ablation}
\end{minipage}
\end{figure}

To isolate the contribution of different context components, we compared performance when providing the LLM with only structured entity information ("Entity-Only") versus the full context including source text ("Entity+Chunk"). As shown in Figure~\ref{fig:context_ablation}, the results are stark. The "Entity-Only" method performs poorly, confirming that while our entity-centric search is powerful for \textit{identifying} the correct reasoning path, the LLM requires the original, grounded text from the source chunks to understand nuance and synthesize a high-quality answer. This validates that our engrams act as a precise \textbf{index}, but the text chunks provide the essential \textbf{content} for final reasoning.

\subsubsection{Impact of Retrieval Depth.}
We analyzed how the associative search depth affects performance, using HotpotQA as a representative case. Table~\ref{tab:depth_ablation} shows that performance peaks at a depth of 2. This confirms that multi-hop traversal is essential, as a depth of 0 or 1 is insufficient to connect the "bridge" entities required by the dataset. Crucially, the average processing time remains stable across all depths, demonstrating that deeper, more accurate retrieval is a quality-control knob with negligible latency cost.

\begin{table}[ht]
\centering
\caption{Ablation on retrieval depth for EcphoryRAG on the HotpotQA dataset. Performance peaks at depth=2.}
\label{tab:depth_ablation}

\begin{tabular}{@{}lccc@{}}
\toprule
\textbf{Retrieval Depth} & \textbf{EM} (\(\uparrow\)) & \textbf{F1-Score} (\(\uparrow\)) & \textbf{Avg Time (s)} (\(\downarrow\)) \\
\midrule
depth 0 & 0.714 & 0.8085 & \textbf{6.612} \\
depth 1 & 0.702 & 0.8032 & 6.690 \\
\textbf{depth 2} & \textbf{0.722} & \textbf{0.8143} & 6.734 \\
depth 3 & 0.698 & 0.7943 & 6.816 \\
\bottomrule
\end{tabular}%

\end{table}

\subsubsection{Impact of Context Size (k).}
Figure~\ref{fig:k_ablation} reveals that the optimal number of retrieved passages (\(k\)) is highly dependent on the dataset's characteristics. Noise-sensitive datasets like 2Wiki benefit from a small, focused context (\(k=20\)), while more noise-tolerant datasets like HotpotQA and MuSiQue require a larger context (\(k=40\) to \(k=80\)) to gather all necessary dispersed evidence. This highlights the importance of tuning \(k\) to balance information richness against the "lost-in-the-middle" problem.


\section{Conclusion}

In this paper, we introduced EcphoryRAG, a novel framework that operationalizes the cognitive principle of memory ecphory for complex, multi-hop question answering. By integrating a cue-driven associative search with a highly efficient entity-centric indexing strategy, our framework sets a new state-of-the-art on challenging QA benchmarks. The results validate that our cognitively inspired design leads not only to superior reasoning accuracy but also to a more practical and scalable architecture for structured RAG systems.

\subsection{Limitations and Future Work}
While our results are promising, the quality of the initial entity extraction remains a critical dependency. Looking forward, the principles of EcphoryRAG pave the way for exciting advancements in lifelong learning and autonomous agents. Future work will explore three main directions:

\begin{enumerate}[label=(\arabic*)]
    \item \textbf{Towards True Continual Learning:} Our framework's key advantage is its highly efficient indexing process. This makes it feasible to dynamically update the knowledge graph as new information arrives, a critical requirement for lifelong learning. Future work will focus on developing memory consolidation mechanisms—analogous to sleep—that can abstract and integrate new knowledge without catastrophically forgetting old information.

    \item \textbf{Advanced Memory for Autonomous Agents:} We aim to extend the EcphoryRAG memory system for autonomous agents. This involves creating a more sophisticated graph. The retrieval mechanism can be adapted to populate a dynamic "working memory" buffer, conditioned on the agent's current task and environment.

    \item \textbf{Goal-Oriented Retrieval:} The "cue" in our current system is derived from the user's query. A next-generation system would use a more complex cue, combining the user's instruction with the agent's internal state, goals, and environmental feedback. This would enable the agent to retrieve information that is not just relevant to the immediate question, but also strategically aligned with its long-term objectives.
\end{enumerate}

Ultimately, EcphoryRAG represents a step towards building more cognitively plausible AI. By grounding our engineering solutions in the time-tested principles of human memory, we aim to contribute to a new generation of systems capable of robust, scalable, and continuous learning.

\bibliographystyle{plainnat}
\bibliography{references}

\end{document}